\documentclass[10pt,twocolumn,letterpaper]{article}
\usepackage[pagenumbers]{cvpr}
\usepackage{multirow}
\usepackage{array}
\usepackage{bm}
\usepackage{microtype}
\usepackage{placeins}
\definecolor{gaincolor}{HTML}{2E8B57}
\definecolor{cvprblue}{rgb}{0.21,0.49,0.74}
\usepackage[breaklinks,colorlinks,allcolors=cvprblue]{hyperref}
\usepackage[capitalize]{cleveref}

\crefname{figure}{Fig.}{Fig.}
\Crefname{figure}{Fig.}{Fig.}
\graphicspath{{figures/}}
\newcommand{\norm}[1]{\left\lVert #1\right\rVert_2}
\newcommand{\ind}{\mathbb{I}}
\def\confName{CVPR}
\def\confYear{2027}

\title{Training-Free Building Change Detection with Example-Induced Discriminants}
\author{
Zhi Lu\\
{\small University of Electronic Science}\\
{\small and Technology of China}\\
{\small\ttfamily\href{mailto:zhilu@uestc.edu.cn}{zhilu@uestc.edu.cn}}
\and
Bin Hong\\
{\small Southwest Jiaotong}\\
{\small University}\\
{\small\ttfamily\href{mailto:hb3258828525@my.swjtu.edu.cn}{hb3258828525@my.swjtu.edu.cn}}
\and
Zhenming Peng\\
{\small University of Electronic Science}\\
{\small and Technology of China}\\
{\small\ttfamily\href{mailto:zmpeng@uestc.edu.cn}{zmpeng@uestc.edu.cn}}
}

\begin{document}
\maketitle

\begin{abstract}
Building change detection (BCD) relies on dense annotations that are costly to collect across regions. A few annotated examples offer a more economical way to specify the target, but change maps provide incomplete building supervision: they label only changed buildings and leave the building-containing date unspecified. Our key observation is that these annotations nevertheless reveal features that recognize buildings in other images, including unchanged ones. We present Linear INduced Discriminants (LIND), a training-free framework that turns regional feature contrasts into a reusable linear building classifier. Using a single frozen vision transformer, it unifies building recognition and temporal comparison in the same feature space. This approach requires neither text prompts nor an auxiliary segmentation model. Only one parameter is tuned on a single validation dataset and then kept fixed across datasets and shot settings. Extensive experiments demonstrate strong detection accuracy, cross-dataset transfer, and efficient inference, with extensions to other semantic classes and broader land-cover changes. Change annotations thus provide a compact visual specification for detection in new scenes. Code: \url{https://github.com/lzcn/lind}.
\end{abstract}


\section{Introduction}
\label{sec:intro}

Building change detection (BCD) identifies construction and demolition from images acquired at different times, supporting urban monitoring, infrastructure management, and disaster assessment~\cite{SpatialTemporalAttentionBasedMethodChen20,OvercomingUncertaintyChallengesLi25}. Supervised detectors achieve strong performance~\cite{RemoteSensingImageChen22,ChangerFeatureInteractionFang23,TransformerbasedSiameseNetworkBandara22,SNUNetCDDenselyConnectedFang22}, but obtaining their labels requires comparing both dates and delineating genuine changes amid appearance variation. Extending monitoring to new regions also introduces differences in building appearance and seasonal conditions that challenge generalization~\cite{CrossDomainChangeDetectionSong23,CSDACDDomainAdaptiveChangeHuang24}. An effective detector should therefore make economical use of annotations while remaining applicable beyond the scenes from which they were collected.

Foundation models offer an opportunity to meet these requirements by reusing representations learned through large-scale pretraining~\cite{LearningTransferableVisualRadford21,DINOv2LearningRobustOquab24,RemoteCLIPVisionLanguageLiu24,SegmentAnythingKirillov23}. Adapting the Segment Anything Model (SAM) to change detection enables training-free change localization~\cite{SegmentAnyChangeZheng24,SegmentChangeModelTan24}. Category-specific detection additionally requires relating temporal differences to the target concept. Open-vocabulary change detection (OVCD) addresses this by using language to specify the target category~\cite{DynamicEarthHowFarLi26,AdaptOVCDTrainingfreeOpenvocabularyDou26,ReAOVCDTrainingFreeOpenVocabularyZhu26}. This flexibility depends on grounding a category description in the image. For buildings, a description can convey the object category while leaving roof appearance and spatial extent underspecified. Prompt and refinement settings suited to one region may also transfer poorly to another. Many systems additionally combine recognition, segmentation, and temporal comparison through intermediate masks or category assignments, allowing errors to propagate between stages~\cite{AdaptOVCDTrainingfreeOpenvocabularyDou26,OpenDPROpenVocabularyChangeGuo26}. These limitations motivate a more direct connection between specifying the target and detecting its changes.

\begin{figure}[!t]
\centering
\includegraphics[width=\linewidth]{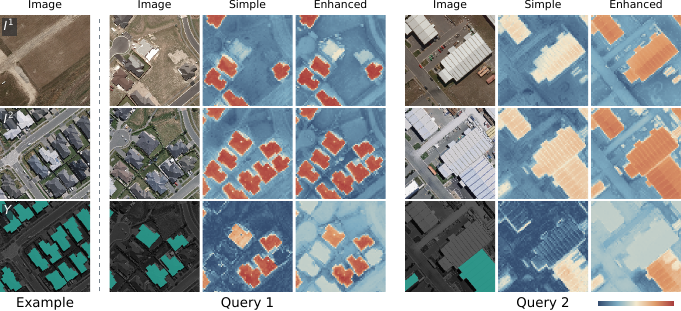}
\caption{Example-based building change detection using interior-prototype matching (Simple) and our method (Enhanced). Corresponding maps share color scales across queries.}
\label{fig:motivation-overview}
\end{figure}

Annotated examples provide a concrete visual specification through the target's appearance and extent. Visual prompting and example-conditioned segmentation demonstrate the value of this interface~\cite{ImagesSpeakImagesWang23,INSID3TrainingfreeIncontextCuttano26,MatcherSegmentAnythingLiu23}. Few-shot change detection also uses annotated example pairs, but trains a task-specific network to interpret them~\cite{FewShotChangeDetectionCao25}. We study how these examples can define a detector directly in frozen features, without task-specific training. Our starting observation is that features pooled from an annotated change region respond to buildings in other images, including unchanged ones, as shown in Fig.~\ref{fig:motivation-overview}. A change annotation therefore exposes reusable building information beyond the particular transition it labels. This motivates constructing a building classifier from change examples and applying it to new scenes.

Extracting this information requires resolving the ambiguity of change annotations. A change map marks where a building appeared or disappeared but does not identify which date contains the building. The same pixels may be a building in one date and its surroundings in the other, and unchanged buildings outside the mask remain mixed with the background. The marked region and its complement therefore cannot serve as complete foreground and background labels. The central challenge is to recover a building-separating direction from these incomplete temporal annotations.

We present Linear INduced Discriminants (LIND), a training-free BCD framework conditioned on change-annotated example pairs. Our key idea is to turn the contrast between an annotated region and its surroundings into a reusable building discriminant. Regional contrasts provide both a cue to the building-containing date and a direction that separates buildings from their context. This construction uses incomplete change annotations to specify a linear decision rule directly, without learning a prediction head or requiring complete building masks. The resulting classifier can recognize both changed and unchanged buildings, allowing temporal comparison to focus on the target category.

Recognizable building responses must also support consistent change decisions across scenes. Response levels can vary with the example and query, while small registration errors can compare different structures across dates. We formulate change scoring as a comparison of building responses and feature dissimilarity, accounting for response variation and local misalignment within each query pair~\cite{LearningDetectChangesMesquita22,DiffRegCDIntegratedRegistrationMadani26}. In Fig.~\ref{fig:motivation-overview}, the final scores suppress unchanged buildings across the two queries while emphasizing changed regions. Building recognition and change scoring thus operate within the same frozen feature space.

Our method requires no text prompts, auxiliary segmentation models, or clustering. Its only validation-selected parameter is a decision margin, fixed across datasets and shot settings. Experiments demonstrate accurate detection, cross-dataset transfer, and efficient inference, while building recognition confirms that change annotations provide information beyond the marked transitions.

Our contributions are threefold:
\begin{itemize}
    \item To the best of our knowledge, we introduce the first training-free BCD framework conditioned directly on change-annotated image pairs, using examples to specify the detector without task-specific learning.
    \item We propose a simple yet effective algorithm that resolves temporal ambiguity and incomplete building supervision through regional feature contrasts, producing a linear discriminant for building recognition and change detection.
    \item Extensive experiments demonstrate state-of-the-art performance among the compared training-free methods on LEVIR-CD and WHU-CD, strong cross-dataset transfer, and fast, memory-efficient inference with one validation-selected margin.
\end{itemize}

\begin{figure*}[!t]
\centering
\includegraphics[width=\linewidth]{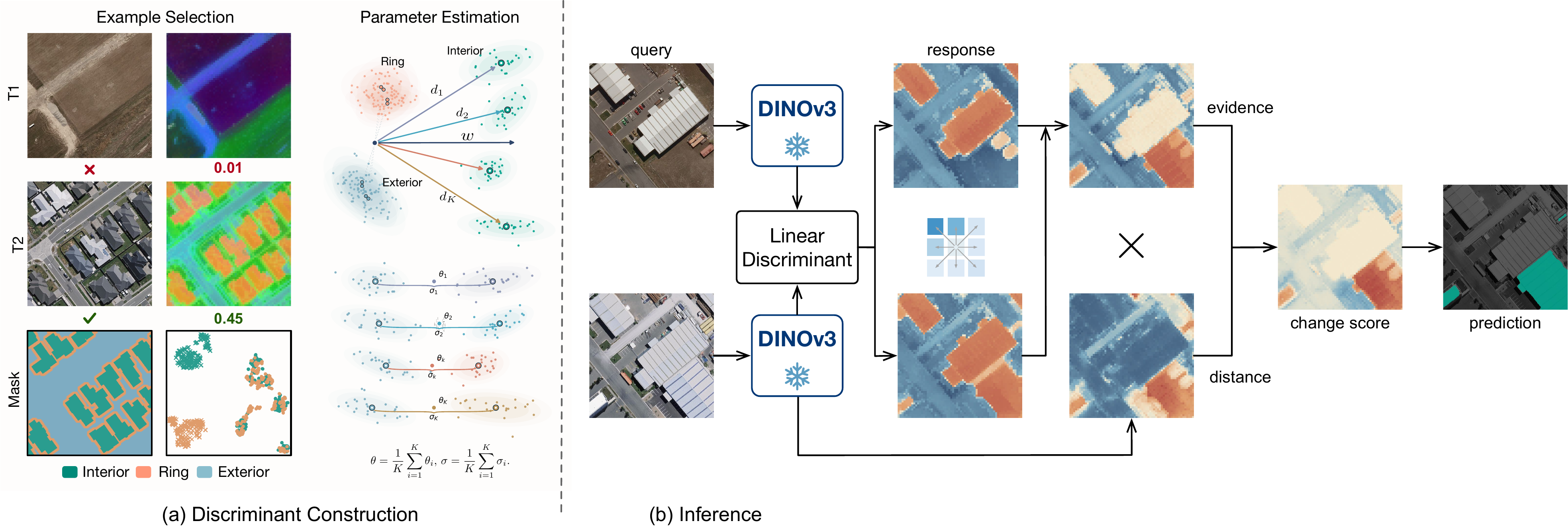}
\caption{Overview of our method. Example-induced discriminants and locally matched query responses produce change scores. Point clouds are schematic.}
\label{fig:framework}
\end{figure*}

\section{Related Work}

\subsection{Supervised Change Detection}
Classical change detectors compare handcrafted spectral or statistical features~\cite{ChangeVectorAnalysisMalila80,RegularizedIterativelyReweightedNielsen07,UnsupervisedChangeDetectionCelik09}. They require no training but depend on threshold selection and are sensitive to radiometric variation. Supervised models learn temporal representations from annotated image pairs for change localization~\cite{ChangeDetectionMethodsCheng24} and description~\cite{HIMECDirectionalChangeAshraf26}. Siamese networks encode both dates with shared weights~\cite{FullyConvolutionalSiameseCayeDaudt18,SNUNetCDDenselyConnectedFang22}, while attention and transformer blocks model spatial and temporal interactions~\cite{SpatialTemporalAttentionBasedMethodChen20,RemoteSensingImageChen22,TransformerbasedSiameseNetworkBandara22,ChangerFeatureInteractionFang23,ChangeMambaRemoteSensingChen24}. Boundary refinement improves spatial detail~\cite{EdgeRefNetEdgeguidedRefinementHussin26}, and semantic variants predict land-cover transitions~\cite{ChangeMaskDeepMultitaskZheng22}.

These models achieve strong in-domain accuracy but depend on the training distribution. Domain adaptation and self-training reduce this dependence by adapting to target-related data or exploiting unlabeled imagery~\cite{CrossDomainChangeDetectionSong23,CSDACDDomainAdaptiveChangeHuang24,AWDAAdversarialWeightedZhang25,SpatiallySelectiveSelfTrainingHussin26}. Few-shot and partial-supervision methods further reduce annotation requirements~\cite{FewShotChangeDetectionCao25,OCAObjectbasedChangeZhang25,FocusedInformationLearningAhn24}. These methods retain task-specific learning.

\subsection{Foundation-Model Change Detection}
Frozen foundation models provide dense representations for temporal comparison without task-specific training. AnyChange uses SAM features and bitemporal latent matching to produce class-agnostic change maps, with point queries specifying objects of interest~\cite{SegmentAnyChangeZheng24}. UCD-SCM applies a frozen segmentation model to building change detection and adds feature recalibration and semantic attention~\cite{SegmentChangeModelTan24}. Other pipelines use SAM proposals or reconstruction to localize changes~\cite{MergeSAMUnsupervisedChangeHu25,UnsupervisedChangeDetectionNoh24}. Feature dissimilarity provides a general change signal, while category-specific detection additionally requires identifying the target concept.

Open-vocabulary change detection supplies this category information through language and pretrained recognition models~\cite{DynamicEarthHowFarLi26,UniVCDNewMethodZhu25}. AdaptOVCD and ReA-OVCD combine semantic recognition with segmentation and temporal comparison~\cite{AdaptOVCDTrainingfreeOpenvocabularyDou26,ReAOVCDTrainingFreeOpenVocabularyZhu26}. MemOVCD and CoRegOVCD incorporate temporal memory and consistency or geometric cues~\cite{MemOVCDTrainingFreeOpenVocabularyKuang26,CoRegOVCDConsistencyregularizedOpenvocabularyTang26}. OmniOVCD combines semantic and instance outputs within SAM~3~\cite{OmniOVCDStreamliningOpenVocabularyZhang26}. OpenDPR retrieves proposals against diffusion-generated visual prototypes~\cite{OpenDPROpenVocabularyChangeGuo26}, and Seg2Change adapts an open-vocabulary segmentation model to bitemporal comparison~\cite{Seg2ChangeAdaptingOpenVocabularySu26}. Target specification and temporal comparison are linked through masks, category assignments, or prototypes. Robustness depends on both semantic grounding and the consistency of these intermediate outputs across dates.

\subsection{In-Context Segmentation}
In-context segmentation uses annotated image--mask examples to specify the target in a query image. Examples specify both target appearance and spatial extent. Few-shot methods learn prototype matching or feature interactions~\cite{PFENetPriorGuidedTian22}, while pretrained models enable inference from examples without task-specific training~\cite{PersonalizeSegmentAnythingZhang24,SegGPTSegmentingWang23,MatcherSegmentAnythingLiu23}. Recent approaches operate directly on frozen features. INSID3 groups features into regions~\cite{INSID3TrainingfreeIncontextCuttano26}, REBASE uses reference and background subspaces~\cite{REBASEReferenceBackgroundSubspaceGopal26}, FSSDINO constructs class prototypes~\cite{SemanticSelectionGapZakir26}, and FROST compares foreground and background densities~\cite{FROSTTrainingFreePark26}. Their example masks generally identify the target in a particular image. Change maps introduce an additional ambiguity by marking a transition without identifying the building-containing date or labeling unchanged buildings. Our method uses regional contrasts to recover a building discriminant from these annotations and compare its responses across dates.

\section{Method}
\label{sec:method}
Our key idea is to construct a linear building discriminant directly from annotated examples in a frozen feature space, without task-specific training. Fig.~\ref{fig:framework} illustrates the two stages of our method. Regional feature contrasts within the examples define the discriminant. Its responses on a query pair, weighted by temporal feature dissimilarity, identify building changes.

\textbf{Problem formulation.}
Given a bitemporal query pair $(I_q^1,I_q^2)$, our goal is to predict its binary change map $Y$. Let $B^1$ and $B^2$ denote the underlying building masks at the two dates. The target is defined pixel-wise as
\begin{equation}
 Y=\ind[B^1\ne B^2].
\label{eq:change-definition}
\end{equation}

Prediction uses the query images and $K$ annotated example pairs, with the pretrained encoder kept frozen:
\begin{equation}
 \mathcal S_c=\{(I_k^1,I_k^2,Y_k)\}_{k=1}^{K}.
\label{eq:example-set}
\end{equation}
When building masks are available for the examples, we instead use
\begin{equation}
 \mathcal S_b=\{(I_k^1,I_k^2,B_k^1,B_k^2)\}_{k=1}^{K}.
\label{eq:building-example-set}
\end{equation}

\textbf{Notation.}
Let $X=\{\bm{x}_u\}_{u\in\mathcal G}$ be a feature map on token grid $\mathcal G$, where $\bm{x}_u\in\mathbb R^d$ is the feature at position $u$. We denote unit-length normalization by
\begin{equation}
 \mathcal N(\bm{v})={\bm{v}}/{\norm{\bm{v}}}.
\label{eq:feature-normalization}
\end{equation}
For a region $S\subseteq\mathcal G$, its mean feature and normalized mean are
\begin{equation}
 m(S;X)=\frac{1}{|S|}\sum_{u\in S}\bm{x}_u,\,
 \pi(S;X)=\mathcal N\!\left(m(S;X)\right).
\label{eq:region-means}
\end{equation}


\subsection{Feature Encoding and Mask Projection}
We represent images and example annotations as features and masks on the token grid $\mathcal G$.

\textbf{Image encoder.}
Let $\bm f_u^t\in\mathbb R^d$ denote the normalized DINOv3 ViT-L/16 feature at position $u$ in date $t$~\cite{DINOv3Simeoni26}. Following INSID3~\cite{INSID3TrainingfreeIncontextCuttano26}, we reduce positional bias by projecting out a subspace estimated from a black image. We normalize its token features, center them across positions, and take the leading $r=500$ left singular vectors as $U_r$. Building recognition and temporal comparison then use the projected features
\begin{equation}
 \bm{x}_u^t=\mathcal N\!\left((I-U_rU_r^\top)\bm{f}_u^t\right).
\label{eq:encoded-feature}
\end{equation}

\textbf{Mask projection.}
We project the example change map $Y$ or building mask $B^t$ onto $\mathcal G$ by area averaging. Let $M(u)$ denote the annotated fraction of token $u$. Fully covered and uncovered tokens define the interior and exterior sets:
\begin{equation}
\Omega^+ = \{u:M(u)=1\},\,
\Omega^- = \{u:M(u)=0\}.
\end{equation}
As illustrated in Fig.~\ref{fig:framework}, the ring $\Gamma\subseteq\Omega^-$ contains exterior tokens immediately adjacent to the annotated region and provides a local exterior reference.

\subsection{Discriminant Construction}
The examples define a linear building response that can be evaluated directly on query features:
\begin{equation}
 z(\bm{x})=\frac{\bm{w}^\top\bm{x}-\theta}{\sigma},
\label{eq:building-score}
\end{equation}
where $\bm{w}$ is the discriminant direction, $\theta$ is the decision boundary, and $\sigma$ is the response scale. We compute all three quantities from regional prototypes without training.

\textbf{Region prototypes.}
The interior, exterior, and ring prototypes of feature map $X$ are
\begin{equation}
 \bm{p} = m(\Omega^+;X),\,
 \bm{g} = m(\Omega^-;X),\,
 \bm{\ell} = m(\Gamma;X).
\label{eq:region-prototypes}
\end{equation}
Building masks specify the building regions in each date. A change map, however, leaves the building-containing date unspecified. We use interior--ring separation as a cue, selecting the date with the larger cosine distance between these prototypes:
\begin{equation}
 t_k^*=\arg\max_{t\in\{1,2\}}\left[1-\langle\pi(\Omega_k^+;X_k^t),\pi(\Gamma_k;X_k^t)\rangle\right].
\label{eq:polarity}
\end{equation}
This criterion favors the date in which the annotated region is distinct from its immediate surroundings, as illustrated in Fig.~\ref{fig:framework}.

\textbf{Discriminant direction.}
The cross-image building responses in Fig.~\ref{fig:motivation-overview} motivate using regional prototypes for building recognition. To distinguish buildings from their surroundings, we define the direction for example $k$ as
\begin{equation}
 \bm{d}_k=\bm{p}_k-\frac12(\bm{g}_k+\bm{\ell}_k).
\label{eq:example-contrast}
\end{equation}
Change maps leave unchanged buildings unmarked in the exterior, so its mean can include building features. Balancing this mean with the ring prototype emphasizes the immediate surroundings of the annotated buildings. Averaging the directions from $K$ examples and normalizing yields the unit direction:
\begin{equation}
 \bm{w}=\mathcal N\!\left(\frac1K\sum_{k=1}^{K}\bm{d}_k\right).
\label{eq:axis}
\end{equation}

\textbf{Decision boundary and scale.}
For each example $k$, we set the decision boundary to the midpoint of the interior and exterior projections and the response scale to the distance between them:
\begin{equation}
 \theta_k=\frac{1}{2}\left(\bm{w}^\top\bm{p}_k+\bm{w}^\top\bm{g}_k\right),\,
 \sigma_k=|\bm{w}^\top(\bm{p}_k-\bm{g}_k)|.
\label{eq:reference-boundary}
\end{equation}
We average over the $K$ examples to obtain the boundary and scale:
\begin{equation}
  \theta=\frac1K\sum_{k=1}^{K}\theta_k,\, \sigma=\frac1K\sum_{k=1}^{K}\sigma_k.
\label{eq:boundary}
\end{equation}
Dividing by $\sigma$ expresses responses relative to the observed interior--exterior separation, providing a common scale for the decision margin.

\subsection{Change Scoring}
Given a query pair $(I_q^1,I_q^2)$, we apply the linear classifier to both feature maps to obtain responses $z_u^t=z(\bm x_u^t)$, using $\theta_q$ in place of $\theta$ as defined below. We define semantic change evidence at position $u$ as
\begin{equation}
 e(u)=\max\left\{z_u^1-\max(z_u^2,0),\; z_u^2-\max(z_u^1,0)\right\}.
\label{eq:crossing}
\end{equation}
Evidence is positive when the building response in one date exceeds both zero and the response in the other date. Feature dissimilarity further downweights evidence between similar tokens:
\begin{equation}
 \delta(u)=1-\langle\bm{x}_u^1,\bm{x}_u^2\rangle.
\label{eq:aligned-distance}
\end{equation}
The resulting change score is
\begin{equation}
 c(u)=e(u)\delta(u).
\label{eq:aligned-score}
\end{equation}

\textbf{Registration compensation.}
Remote sensing image pairs may contain slight registration errors~\cite{LearningDetectChangesMesquita22,DiffRegCDIntegratedRegistrationMadani26}. We therefore match each token to the most similar feature in the other date within the local neighborhood
\begin{equation}
 \mathcal W(u)=\{v\in\mathcal G:\|u-v\|_\infty\le1\}.
\label{eq:neighborhood}
\end{equation}
Within $\mathcal W(u)$, $v$ maximizes cosine similarity to $\bm{x}_u^1$ in date 2, and $w$ maximizes cosine similarity to $\bm{x}_u^2$ in date 1. We then compute the evidence and feature dissimilarity as
\begin{equation}
 e(u)=\max\{z_u^1-\max(z_v^2,0),\,z_u^2-\max(z_w^1,0)\},
\label{eq:matched}
\end{equation}
\begin{equation}
 \delta(u)=\min\{1-\langle\bm{x}_u^1,\bm{x}_v^2\rangle,\,1-\langle\bm{x}_u^2,\bm{x}_w^1\rangle\}.
\label{eq:matched-distance}
\end{equation}
We substitute these matched quantities into Eq.~\eqref{eq:aligned-score} to obtain the change score.

\textbf{Boundary calibration.}
To account for query-specific response offsets, we split the projections $r_u^t=\bm w^\top\bm x_u^t$ above and below the example boundary $\theta$ from Eq.~\eqref{eq:boundary}. Let $\bar r_+$ and $\bar r_-$ be the respective means across both dates. We average their midpoint with the example boundary:
\begin{equation}
 \theta_q=\frac12(\theta+m_q),\ \text{where }m_q=\frac12(\bar r_++\bar r_-).
\label{eq:calibration}
\end{equation}
This update uses only the current query pair and keeps $\bm w$ and $\sigma$ unchanged.

Finally, we bilinearly upsample the token-level change scores $c$ to image resolution and apply the decision margin:
\begin{equation}
 \widehat Y=\ind[\mathcal B(c)>\tau],
\label{eq:pixel-prediction}
\end{equation}
where $\mathcal B$ denotes bilinear interpolation. The margin $\tau$ is the only tuned parameter in our method, selected on the validation split of one dataset and used unchanged across all datasets.

\section{Experiments}
\label{sec:experiments}
\label{sec:full-results}
\subsection{Experimental Setup}

\textbf{Datasets.}
We evaluate building change detection on LEVIR-CD~\cite{SpatialTemporalAttentionBasedMethodChen20}, WHU-CD~\cite{FullyConvolutionalNetworksJi19}, and S2Looking~\cite{S2LookingSatelliteSidelookingShen21}, with SECOND~\cite{AsymmetricSiameseNetworksYang22} and DSIFN-CD~\cite{DeeplySupervisedImageZhang20} extending the evaluation to other semantic classes and broader land-cover changes. LEVIR-CD contains 637 image pairs at $1024\times1024$ pixels, officially split into 445 training, 64 validation, and 128 test pairs. WHU-CD provides 1,260 training and 690 test pairs at $512\times512$, with no official validation split. S2Looking contains 5,000 side-looking satellite pairs at $1024\times1024$, officially split into 3,500 training, 500 validation, and 1,000 test pairs. SECOND labels six land-cover classes in 4,662 pairs at $512\times512$, with 2,968 training and 1,694 test pairs and no validation split. DSIFN-CD labels general land-cover changes in 3,988 pairs at $512\times512$, officially split into 3,600 training, 340 validation, and 48 test pairs. We remove duplicates from its augmented training and validation sets.



\textbf{Evaluation protocol and metrics.}
To estimate performance across example selections, we independently sample $K$ training example pairs for each query unless stated otherwise. We call this the $K$-shot setting and evaluate $K\in\{1,5,10\}$. We report foreground IoU and F1 as percentages, computed from pooled pixel counts.

\textbf{Implementation details.}
We use $256\times256$ crops for LEVIR-CD and $512\times512$ crops for the other datasets. Each crop is upsampled to $1024\times1024$ and encoded by frozen DINOv3 ViT-L/16 into a $64\times64$ token grid using final-layer features. We select $\tau=0.095$ on 10-shot LEVIR-CD validation and keep it fixed across datasets and shot settings.

\textbf{Baselines.}
We compare conventional, frozen-feature, and open-vocabulary change detectors, together with example-conditioned segmentation methods INSID3~\cite{INSID3TrainingfreeIncontextCuttano26} and FROST~\cite{FROSTTrainingFreePark26}. For the latter two, we segment both query dates and take the XOR of their masks. BIT~\cite{RemoteSensingImageChen22} serves as a supervised reference.

\subsection{Building Change Detection Performance}

\begin{table}[!t]
\centering
\normalsize
\caption{Building change detection on LEVIR-CD and WHU-CD.}
\label{tab:main-results}
\begin{tabular}{lcccc}
\hline
\multirow{2}{*}{Method} & \multicolumn{2}{c}{LEVIR-CD} & \multicolumn{2}{c}{WHU-CD} \\
\cline{2-3}\cline{4-5}
& IoU & F1 & IoU & F1 \\
\hline
CVA~\cite{ChangeVectorAnalysisMalila80} & 5.41 & 10.26 & 3.71 & 7.16 \\
IRMAD~\cite{RegularizedIterativelyReweightedNielsen07} & 6.72 & 12.59 & 3.31 & 6.41 \\
PCA-K~\cite{UnsupervisedChangeDetectionCelik09} & 5.36 & 10.18 & 6.64 & 12.46 \\
ISFA~\cite{PostclassificationChangeDetectionWu17} & 5.36 & 10.17 & 3.82 & 7.36 \\
DSFA~\cite{UnsupervisedDeepSlowDu19} & 7.92 & 14.67 & 5.91 & 11.15 \\
DCVA~\cite{UnsupervisedDeepChangeSaha19} & 13.28 & 24.50 & 15.62 & 28.36 \\
\hline
AnyChange~\cite{SegmentAnyChangeZheng24} & 7.85 & 14.56 & 7.30 & 13.59 \\
UCD-SCM~\cite{SegmentChangeModelTan24} & 19.15 & 32.15 & 21.58 & 35.50 \\
IMC~\cite{DynamicEarthHowFarLi26} & 49.68 & 65.78 & 53.76 & 69.68 \\
MCI~\cite{DynamicEarthHowFarLi26} & 36.08 & 53.03 & 39.06 & 56.17 \\
OpenDPR~\cite{OpenDPROpenVocabularyChangeGuo26} & 44.82 & 61.90 & 64.83 & 78.67 \\
  AdaptOVCD~\cite{AdaptOVCDTrainingfreeOpenvocabularyDou26} & 51.52 & 68.00 & 61.99 & 76.54 \\
OmniOVCD~\cite{OmniOVCDStreamliningOpenVocabularyZhang26} & 67.80 & 80.81 & 75.86 & 86.27 \\
MemOVCD~\cite{MemOVCDTrainingFreeOpenVocabularyKuang26} & 72.50 & 84.10 & 71.03 & 83.06 \\
CoRegOVCD~\cite{CoRegOVCDConsistencyregularizedOpenvocabularyTang26} & 56.92 & 72.55 & 56.09 & 71.87 \\
ReA-OVCD~\cite{ReAOVCDTrainingFreeOpenVocabularyZhu26} & 73.21 & 84.53 & \underline{82.35} & \underline{90.32} \\
INSID3~\cite{INSID3TrainingfreeIncontextCuttano26} & 60.27 & 75.21 & 42.79 & 59.93 \\
FROST~\cite{FROSTTrainingFreePark26} & 58.31 & 73.66 & 34.10 & 50.85 \\
\hline
Ours (1-shot) & 64.45 & 78.38 & 56.83 & 72.48 \\
Ours (5-shot) & \underline{73.64} & \underline{84.82} & 80.58 & 89.25 \\
Ours (10-shot) & \textbf{74.17} & \textbf{85.17} & \textbf{82.84} & \textbf{90.62} \\
\hline
BIT~\cite{RemoteSensingImageChen22} & 79.79 & 88.76 & 84.55 & 91.63 \\
\hline
\end{tabular}
\end{table}

\begin{figure*}[!t]
\centering
\includegraphics[width=\linewidth]{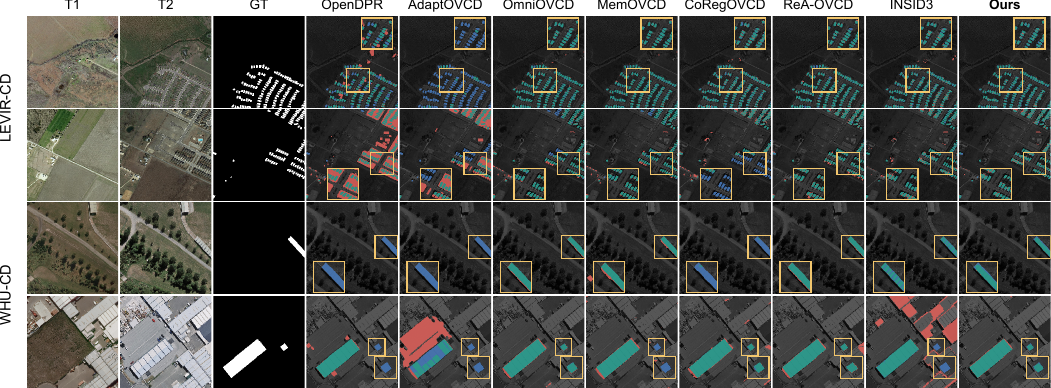}
\caption{Predictions on LEVIR-CD and WHU-CD with enlarged details. \textcolor[HTML]{2A9D8F}{\rule{0.6em}{0.6em}} TP, \textcolor[HTML]{D95F59}{\rule{0.6em}{0.6em}} FP, \textcolor[HTML]{4575B4}{\rule{0.6em}{0.6em}} FN.}
\label{fig:qualitative}
\end{figure*}

We evaluate whether a few change-annotated examples can define an effective building change detector without training. Table~\ref{tab:main-results} compares performance on LEVIR-CD and WHU-CD. In the 5-shot setting, our method outperforms ReA-OVCD, the strongest OVCD baseline, on LEVIR-CD and approaches its performance on WHU-CD. In the 10-shot setting, it achieves the highest IoU and F1 among the compared training-free methods on both datasets. With the same frozen encoder and 10-shot examples, our method also outperforms the segmentation-based comparisons INSID3 and FROST. These comparisons favor deriving change evidence directly from building responses.

The largest gains occur from 1-shot to 5-shot, with smaller gains from 5-shot to 10-shot. Multiple examples reduce dependence on a particular annotated transition, with diminishing gains as their number increases.

Fig.~\ref{fig:qualitative} shows these gains in dense layouts and small structures. In the first two rows, our method separates neighboring changed buildings, while several compared methods miss buildings or produce false detections near their boundaries. It also detects the isolated small building in the third row and the small changed structure beside the warehouse in the fourth.

\begin{table}[!t]
\centering
\normalsize
\caption{Annotation comparison on WHU-CD.}
\label{tab:support-interface}
\label{tab:building-segmentation}
\setlength{\tabcolsep}{2pt}
\begin{tabular}{@{}ccccccc|cc@{}}
\hline
\multirow{2}{*}{$K$} & \multicolumn{2}{c}{$\mathcal S_c$} & \multicolumn{2}{c}{$\mathcal S_b$} & \multicolumn{2}{c|}{$\Delta$} & \multicolumn{2}{c}{Building F1} \\
\cline{2-3}\cline{4-5}\cline{6-7}\cline{8-9}
& IoU & F1 & IoU & F1 & IoU & F1 & $\mathcal S_c$ & $\mathcal S_b$ \\
\hline
1 & 56.83 & 72.48 & 71.17 & 83.16 & +14.34 & +10.68 & 81.14 & 93.09 \\
5 & 80.58 & 89.25 & 81.52 & 89.82 & +0.94 & +0.57 & 93.45 & 93.54 \\
10 & 82.84 & 90.62 & 83.12 & 90.78 & +0.28 & +0.17 & 93.71 & 93.54 \\
\hline
\end{tabular}
\end{table}

\begin{table}[!t]
\centering
\normalsize
\caption{Transfer to S2Looking.}
\label{tab:additional-datasets}
\begin{tabular}{lcccc}
\hline
Source & \multicolumn{2}{c}{LEVIR-CD} & \multicolumn{2}{c}{WHU-CD} \\
\cline{2-3}\cline{4-5}
Method & IoU & F1 & IoU & F1 \\
\hline
OpenDPR~\cite{OpenDPROpenVocabularyChangeGuo26} & 23.71 & 38.33 & 6.31 & 11.87 \\
AdaptOVCD~\cite{AdaptOVCDTrainingfreeOpenvocabularyDou26} & 24.38 & 39.20 & 4.72 & 9.01 \\
OmniOVCD~\cite{OmniOVCDStreamliningOpenVocabularyZhang26} & 22.67 & 36.96 & 22.26 & 36.42 \\
MemOVCD~\cite{MemOVCDTrainingFreeOpenVocabularyKuang26} & 17.11 & 29.22 & 11.20 & 20.14 \\
CoRegOVCD~\cite{CoRegOVCDConsistencyregularizedOpenvocabularyTang26} & 18.87 & 31.74 & 3.86 & 7.44 \\
ReA-OVCD~\cite{ReAOVCDTrainingFreeOpenVocabularyZhu26} & 28.40 & 44.23 & 31.28 & 47.65 \\
\hline
\multicolumn{5}{l}{{Source-dataset support}} \\
Ours (1-shot) & 19.10 & 32.08 & 17.00 & 29.06 \\
Ours (5-shot) & 29.16 & 45.15 & 32.78 & 49.37 \\
Ours (10-shot) & \textbf{30.74} & \textbf{47.03} & \textbf{34.64} & \textbf{51.46} \\
\hline
\multicolumn{5}{l}{{Target-dataset support}} \\
Ours (1-shot) & 29.47 & 45.52 & 29.47 & 45.52 \\
Ours (5-shot) & 38.50 & 55.59 & 38.50 & 55.59 \\
Ours (10-shot) & \textbf{38.67} & \textbf{55.77} & \textbf{38.67} & \textbf{55.77} \\
\hline
\end{tabular}
\end{table}

To examine incomplete supervision, Table~\ref{tab:support-interface} compares change maps $\mathcal S_c$ and complete building masks $\mathcal S_b$ on the same WHU-CD examples. Building masks improve detection most in the 1-shot setting, while the two annotation types give comparable performance in the 5-shot and 10-shot settings. Additional change-annotated examples expose more building appearances, narrowing the advantage of complete building masks.

\textbf{Building recognition.}
Building recognition tests whether the induced classifier recovers the target category beyond marked changes. In Table~\ref{tab:building-segmentation}, change maps and complete building masks yield comparable building F1 at 5-shot and 10-shot, despite the incomplete object labels in change maps.

\subsection{Cross-Dataset Transfer}
\label{sec:regional-transfer}
\label{sec:configuration-transfer}
We test whether the example-induced discriminant transfers to new scenes without target-specific tuning. The evaluation covers S2Looking and transfer between LEVIR-CD and WHU-CD. OVCD methods retain their source-dataset settings, while our method uses source training examples and keeps the decision margin unchanged.

S2Looking tests transfer from aerial scenes to side-looking satellite imagery. In Table~\ref{tab:additional-datasets}, our method outperforms the compared OVCD baselines with 5-shot or 10-shot source examples. Target examples further improve performance without changing the decision margin. Source examples retain useful building information across datasets, while target examples better represent local appearance.

\begin{table}[!t]
\centering
\normalsize
\caption{Transfer between LEVIR-CD and WHU-CD.}
\label{tab:cross-dataset}
\setlength{\tabcolsep}{3pt}
\begin{tabular}{lcccc}
\hline
\multirow{2}{*}{Method} & \multicolumn{2}{c}{WHU $\to$ LEVIR} & \multicolumn{2}{c}{LEVIR $\to$ WHU} \\
\cline{2-3}\cline{4-5}
& IoU & F1 & IoU & F1 \\
\hline
AdaptOVCD~\cite{AdaptOVCDTrainingfreeOpenvocabularyDou26} & 8.26 & 15.26 & 45.66 & 62.70 \\
OmniOVCD~\cite{OmniOVCDStreamliningOpenVocabularyZhang26} & 62.06 & 76.59 & 71.47 & 83.36 \\
CoRegOVCD~\cite{CoRegOVCDConsistencyregularizedOpenvocabularyTang26} & 32.18 & 48.70 & 61.12 & 75.87 \\
ReA-OVCD~\cite{ReAOVCDTrainingFreeOpenVocabularyZhu26} & 65.30 & 79.01 & \textbf{78.72} & \textbf{88.10} \\
OpenDPR~\cite{OpenDPROpenVocabularyChangeGuo26} & 35.33 & 52.21 & 56.43 & 72.15 \\
\hline
Ours (1-shot) & 35.24 & 52.11 & 62.37 & 76.82 \\
Ours (5-shot) & \underline{67.98} & \underline{80.94} & 72.95 & 84.36 \\
Ours (10-shot) & \textbf{71.37} & \textbf{83.29} & \underline{75.90} & \underline{86.30} \\
\hline
\end{tabular}
\end{table}

Between LEVIR-CD and WHU-CD, 10-shot source examples retain performance close to the in-domain results in Table~\ref{tab:cross-dataset}. Our method leads when transferring from WHU-CD to LEVIR-CD, while ReA-OVCD remains stronger in the reverse direction.

\begin{figure}[!t]
\centering
\includegraphics[width=\linewidth]{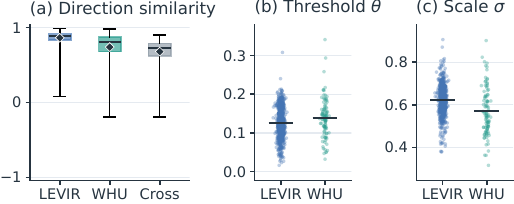}
\caption{Statistics of $\bm w$, $\theta$, and $\sigma$ (10-shot).}
\label{fig:parameter-distributions}
\end{figure}

To examine what transfers, Fig.~\ref{fig:parameter-distributions} compares the induced directions, boundaries, and scales. Directions remain similar across datasets, while boundaries and scales vary, motivating query-boundary calibration.


\subsection{Extension to Other Changes}
\label{sec:second}

To assess applicability beyond building change detection, we evaluate diverse land-cover transitions on DSIFN-CD and changes involving other semantic classes on SECOND.

\textbf{Land-cover changes.}
\label{sec:dsifn-results}
DSIFN-CD combines diverse land-cover transitions in a binary change map. To represent these transitions directly, we construct prototypes from temporal feature differences, denoted as Ours$^\dagger$ in Table~\ref{tab:dsifn-results}. The encoder, example sampling, and decision margin remain unchanged.

\begin{table}[!t]
\centering
\normalsize
\caption{Change detection on DSIFN-CD.}
\label{tab:dsifn-results}
\begin{tabular}{lcccc}
\hline
\multirow{2}{*}{Method} & \multicolumn{2}{c}{Test set} & \multicolumn{2}{c}{Full set} \\
\cline{2-3}\cline{4-5}
& IoU & F1 & IoU & F1 \\
\hline
OpenDPR~\cite{OpenDPROpenVocabularyChangeGuo26} & 37.33 & 54.37 & 22.05 & 36.13 \\
AdaptOVCD~\cite{AdaptOVCDTrainingfreeOpenvocabularyDou26} & 42.21 & 59.37 & 41.69 & 58.84 \\
OmniOVCD~\cite{OmniOVCDStreamliningOpenVocabularyZhang26} & 32.95 & 49.57 & 36.67 & 53.66 \\
MemOVCD~\cite{MemOVCDTrainingFreeOpenVocabularyKuang26} & 33.70 & 50.41 & 25.23 & 40.30 \\
CoRegOVCD~\cite{CoRegOVCDConsistencyregularizedOpenvocabularyTang26} & 32.72 & 49.31 & 43.27 & 60.40 \\
ReA-OVCD~\cite{ReAOVCDTrainingFreeOpenVocabularyZhu26} & \underline{48.74} & \underline{65.54} & 43.87 & 60.99 \\
INSID3~\cite{INSID3TrainingfreeIncontextCuttano26} & 37.86 & 54.93 & 38.21 & 55.30 \\
\hline
Ours & 46.97 & 63.92 & \underline{46.40} & \underline{63.39} \\
Ours$^\dagger$ & \textbf{50.54} & \textbf{67.14} & \textbf{52.10} & \textbf{68.51} \\
\hline
\multicolumn{5}{l}{$^\dagger$Change-vector prototypes.} \\
\end{tabular}
\end{table}

Both evaluations use 10-shot examples. The full-set evaluation pools the deduplicated training, validation, and test splits, drawing each query's examples from the remaining pairs. Change-vector prototypes improve both evaluations in Table~\ref{tab:dsifn-results}, indicating that temporal differences better represent these diverse transitions than a single building direction.

\textbf{Other semantic classes.}
SECOND tests whether examples can specify target categories beyond buildings. It covers six land-cover classes, including buildings, trees, and water. Each class reuses one 10-shot example set for all queries.

\begin{table*}[!t]
\centering
\normalsize
\caption{Class-wise change detection on SECOND. Avg. is the class mean.}
\label{tab:second-classes}
\setlength{\tabcolsep}{3pt}
\begin{tabular}{@{}l*{14}{c}@{}}
\hline
\multirow{2}{*}{Method} & \multicolumn{2}{c}{Building} & \multicolumn{2}{c}{Tree} & \multicolumn{2}{c}{Water} & \multicolumn{2}{c}{Low veg.} & \multicolumn{2}{c}{N.v.g. Surface} & \multicolumn{2}{c}{Playground} & \multicolumn{2}{c}{Avg.} \\
\cline{2-3}\cline{4-5}\cline{6-7}\cline{8-9}\cline{10-11}\cline{12-13}\cline{14-15}
& IoU & F1 & IoU & F1 & IoU & F1 & IoU & F1 & IoU & F1 & IoU & F1 & IoU & F1 \\
\hline
IMC~\cite{DynamicEarthHowFarLi26} & 31.74 & 48.19 & 10.89 & 19.64 & 12.34 & 21.97 & 0.40 & 0.79 & 0.00 & 0.00 & 26.93 & 42.43 & 13.72 & 22.17 \\
MCI~\cite{DynamicEarthHowFarLi26} & 38.65 & 55.75 & 15.38 & 26.66 & 15.39 & 26.68 & 21.17 & 34.95 & 27.91 & 43.65 & 22.39 & 36.59 & 23.48 & 37.38 \\
UniVCD~\cite{UniVCDNewMethodZhu25} & 43.20 & 60.40 & 18.90 & 31.90 & 8.20 & 15.20 & 24.90 & 39.80 & 28.00 & 43.70 & 0.00 & 0.00 & 20.53 & 31.83 \\
OmniOVCD~\cite{OmniOVCDStreamliningOpenVocabularyZhang26} & 45.20 & 62.30 & 16.70 & 28.60 & 21.20 & 35.00 & 24.50 & 39.30 & 27.70 & 43.40 & 27.00 & 42.40 & 27.05 & 41.83 \\
OpenDPR~\cite{OpenDPROpenVocabularyChangeGuo26} & 42.40 & 59.50 & \underline{20.90} & \underline{34.50} & 17.50 & 29.80 & 23.00 & 37.40 & 30.20 & 46.40 & \underline{37.30} & \underline{54.30} & 28.55 & 43.65 \\
AdaptOVCD~\cite{AdaptOVCDTrainingfreeOpenvocabularyDou26} & 44.98 & 62.05 & 12.67 & 22.49 & 23.34 & 37.84 & 21.21 & 34.99 & 33.99 & 50.74 & 29.28 & 45.30 & 27.58 & 42.24 \\
ReA-OVCD~\cite{ReAOVCDTrainingFreeOpenVocabularyZhu26} & \underline{59.12} & \underline{74.31} & 19.40 & 32.48 & 26.09 & 41.39 & 27.95 & 43.69 & \underline{34.80} & \underline{51.63} & 24.74 & 39.67 & \underline{32.02} & \underline{47.19} \\
\hline
Ours ($\mathcal S_c$) & 50.61 & 67.21 & 14.25 & 24.94 & \textbf{33.70} & \textbf{50.41} & \underline{28.08} & \underline{43.85} & 31.39 & 47.78 & 31.04 & 47.38 & 31.51 & 46.93 \\
Ours ($\mathcal S_b$) & \textbf{59.49} & \textbf{74.60} & \textbf{22.72} & \textbf{37.03} & \underline{33.52} & \underline{50.21} & \textbf{29.24} & \textbf{45.24} & \textbf{37.88} & \textbf{54.95} & \textbf{50.79} & \textbf{67.37} & \textbf{38.94} & \textbf{54.90} \\
\hline
\end{tabular}
\end{table*}

Table~\ref{tab:second-classes} compares change map support $\mathcal S_c$ with per-date semantic-mask support $\mathcal S_b$ for the same examples. Here $\mathcal S_b$ identifies the target class separately in each date and excludes unchanged regions with unspecified classes. Mean F1 rises from 46.93\% to 54.90\%, with gains in five of six classes and the largest gain on playgrounds. Explicit class labels help resolve which side of a transition supplies target-class evidence.

\subsection{Ablation Study}
We assess discriminant construction and query scoring by removing six components individually in Table~\ref{tab:matched-readouts}: (a) ring, (b) exterior contrast, (c) scale normalization, (d) registration compensation, (e) boundary calibration, and (f) distance weighting. For (f), we reselect $\tau$ on LEVIR-CD validation.

\begin{table}[!t]
\normalsize
\centering
\caption{Ablation study (F1).}
\label{tab:matched-readouts}
\setlength{\tabcolsep}{4pt}
\begin{tabular}{rccc|ccc}
\hline
& \multicolumn{3}{c}{LEVIR-CD} & \multicolumn{3}{c}{WHU-CD} \\
\cline{2-4}\cline{5-7}
$K$ & 1 & 5 & 10 & 1 & 5 & 10 \\
\hline
Full & \textbf{78.38} & \textbf{84.82} & \textbf{85.17} & \textbf{72.48} & \textbf{89.25} & \textbf{90.62} \\
(a) & 77.69 & 83.44 & 83.69 & \underline{72.15} & \underline{88.76} & 90.07 \\
(b) & 72.22 & 77.93 & 77.47 & 57.41 & 84.73 & 88.60 \\
(c) & 74.95 & 82.63 & 83.11 & 70.32 & 87.55 & 89.69 \\
(d) & \underline{78.04} & 82.82 & 82.91 & 69.57 & 80.97 & 81.36 \\
(e) & 75.59 & \underline{83.79} & \underline{84.06} & 65.36 & 88.04 & \underline{90.41} \\
(f) & 72.72 & 82.42 & 82.81 & 61.45 & 86.02 & 88.66 \\
\hline
\end{tabular}
\end{table}

Interior--exterior contrast is central to discriminant construction: removing it reduces F1. The ring gives modest gains. An interior prototype can retrieve buildings, but exterior contrast makes the response relative to the surrounding scene. The larger loss from removing this contrast, compared with removing the ring, separates the central discriminative mechanism from the refinement provided by local context. When $\sigma$ varies little across examples, its effect can largely be absorbed by retuning $\tau$.

Query scoring improves robustness to response variation and spatial mismatch. Calibration helps most at 1-shot, while registration compensation gives larger gains on WHU-CD. In Table~\ref{tab:registration-stress}, compensation reduces false positives under synthetic token shifts on unchanged validation pairs. Local matches reduce changes falsely attributed to spatial offsets. These effects address different sources of uncertainty: calibration adjusts the response offset inherited from the examples, whereas matching improves temporal correspondence. Their relative gains vary across settings because example variability and spatial misalignment need not be equally pronounced.

\begin{table}[!t]
\centering
\normalsize
\caption{False positives under token shifts with ($\checkmark$) and without ($\times$) registration compensation (FPR, \%).}
\label{tab:registration-stress}
\setlength{\tabcolsep}{2pt}
\begin{tabular}{lccccccccc}
\hline
& \multicolumn{3}{c}{LEVIR-CD} & \multicolumn{3}{c}{WHU-CD} & \multicolumn{3}{c}{S2Looking} \\
\cline{2-4}\cline{5-7}\cline{8-10}
Shift & 0 & 1 & 2 & 0 & 1 & 2 & 0 & 1 & 2 \\
\hline
$\checkmark$ & 0.17 & 0.17 & 0.27 & 0.13 & 0.16 & 1.77 & 0.46 & 0.49 & 1.05 \\
$\times$ & 0.56 & 0.61 & 1.61 & 0.97 & 4.41 & 9.86 & 0.96 & 2.05 & 3.93 \\
\hline
\end{tabular}
\end{table}

%

\begin{figure}[!t]
\centering
\includegraphics[width=\linewidth]{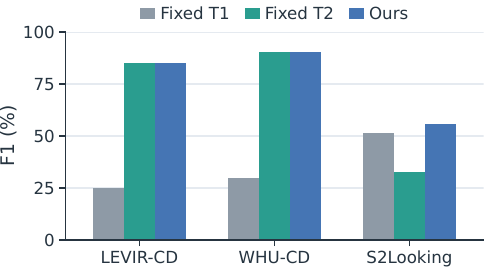}
\caption{Example date selection (10-shot F1, \%).}
\label{fig:image-selection}
\end{figure}

\textbf{Example image selection.}
Fig.~\ref{fig:image-selection} compares example date selection under the final scoring pipeline. Fixed T2 works well on LEVIR-CD and WHU-CD, whereas fixed T1 is stronger on S2Looking. Automatic selection achieves the highest F1 on all three. The preferred fixed date depends on which image contains the annotated buildings. The larger gain on S2Looking shows the value of selecting that date from each example rather than assuming a consistent image order.

\textbf{Backbone comparison.}
We assess the role of frozen representations by varying the encoder and pretraining data in Table~\ref{tab:backbones}. DINOv3-L~\cite{DINOv3Simeoni26} outperforms DINOv2-L~\cite{DINOv2LearningRobustOquab24}, while its satellite-pretrained weights perform worse than its general-image weights. Pretraining on remote sensing imagery alone therefore does not ensure features suited to the induced discriminant. Regional means must provide a distinct building direction that remains useful across images. Matching the pretraining domain is less informative than measuring how the resulting features separate buildings from their surroundings.

\begin{table}[!t]
\centering
\normalsize
\setlength{\tabcolsep}{1.8pt}
\caption{Backbone comparison (F1).}
\label{tab:backbones}
\begin{tabular}{lcccccc}
\hline
\multirow{2}{*}{Backbone} & \multicolumn{3}{c}{LEVIR-CD} & \multicolumn{3}{c}{WHU-CD} \\
\cline{2-4}\cline{5-7}
& $K{=}1$ & $K{=}5$ & $K{=}10$ & $K{=}1$ & $K{=}5$ & $K{=}10$ \\
\hline
DINOv2-L & 65.26 & 74.16 & 75.02 & \underline{62.61} & \underline{81.20} & \underline{83.30} \\
DINOv3-L & \textbf{78.38} & \textbf{84.82} & \textbf{85.17} & \textbf{72.48} & \textbf{89.25} & \textbf{90.62} \\
DINOv3-L$^\dagger$ & \underline{67.87} & \underline{78.68} & \underline{79.77} & 61.13 & 72.66 & 75.76 \\
\hline
\multicolumn{7}{l}{$^\dagger$SAT493M satellite pretraining.} \\
\end{tabular}
\end{table}

\textbf{Computational efficiency.}
To assess computational cost, all methods receive the same $512\times512$ image pair and retain their own model input sizes, with input-to-output latency and peak memory measured on a single RTX~4090 after example preparation. Our method has the lowest latency and memory use in Table~\ref{tab:efficiency}. The same dense features support recognition and temporal comparison, with prototype statistics and local matching completing the inference pipeline.

\begin{table}[!t]
\normalsize
\centering
\caption{Query inference cost. FMs: foundation-model families.}
\label{tab:efficiency}
\setlength{\tabcolsep}{4pt}
\begin{tabular}{lccc}
\hline
Method & FMs & Latency (s) & Memory (GiB) \\
\hline
AdaptOVCD~\cite{AdaptOVCDTrainingfreeOpenvocabularyDou26} & 3 & 4.61 & 10.32 \\
CoRegOVCD~\cite{CoRegOVCDConsistencyregularizedOpenvocabularyTang26} & 2 & 1.51 & 5.55 \\
OmniOVCD~\cite{OmniOVCDStreamliningOpenVocabularyZhang26} & 1 & \underline{0.60} & 4.01 \\
OpenDPR~\cite{OpenDPROpenVocabularyChangeGuo26} & 2 & 0.71 & 6.39 \\
ReA-OVCD~\cite{ReAOVCDTrainingFreeOpenVocabularyZhu26} & 1 & 0.93 & \underline{3.97} \\
\hline
Ours & 1 & \textbf{0.36} & \textbf{1.51} \\
\hline
\end{tabular}
\end{table}

\FloatBarrier

\section{Conclusion}
We presented LIND, a training-free framework that turns change-annotated examples into a linear building discriminant using one frozen vision transformer. Regional feature contrasts recover building information beyond the annotated transitions, enabling recognition and change detection in the same feature space. Extensive experiments demonstrate strong detection and cross-dataset transfer, efficient inference, and applicability to other semantic classes and land-cover changes. With one validation-selected margin, this simple framework offers a practical alternative to task-specific training. Its effectiveness depends on example coverage and frozen features that preserve building identity across scenes.

\label{sec:main-end}
\clearpage
{\small
\bibliographystyle{cvpr}
\bibliography{cvpr}

@misc{AdaptOVCDTrainingfreeOpenvocabularyDou26,
  title = {{{AdaptOVCD}}: Training-Free Open-Vocabulary Remote Sensing Change Detection via Adaptive Information Fusion},
  author = {Dou, Mingyu and Qiu, Shi and Hu, Ming and Chen, Yifan and Ye, Huping and Liao, Xiaohan and Sun, Zhe},
  year = 2026,
  number = {arXiv:2602.06529},
  eprint = {2602.06529},
  primaryclass = {cs.CV},
  doi = {10.48550/arXiv.2602.06529}
}

@article{AsymmetricSiameseNetworksYang22,
  title = {Asymmetric Siamese Networks for Semantic Change Detection in Aerial Images},
  author = {Yang, Kunping and Xia, Gui-Song and Liu, Zicheng and Du, Bo and Yang, Wen and Pelillo, Marcello and Zhang, Liangpei},
  year = 2022,
  journal = {IEEE Transactions on Geoscience and Remote Sensing},
  volume = {60},
  pages = {1--18},
  doi = {10.1109/TGRS.2021.3113912}
}

@article{AWDAAdversarialWeightedZhang25,
  title = {{{AWDA}}: {{Adversarial}} and {{Weighted Domain Adaptation}} for Cross-Dataset Change Detection},
  author = {Zhang, Xueting and Huang, Xin and Li, Jiayi},
  year = 2025,
  journal = {ISPRS Journal of Photogrammetry and Remote Sensing},
  volume = {224},
  pages = {398--409},
  doi = {10.1016/j.isprsjprs.2025.04.008}
}

@article{ChangeDetectionMethodsCheng24,
  title = {Change Detection Methods for Remote Sensing in the Last Decade: A Comprehensive Review},
  author = {Cheng, Guangliang and Huang, Yunmeng and Li, Xiangtai and Lyu, Shuchang and Xu, Zhaoyang and Zhao, Hongbo and Zhao, Qi and Xiang, Shiming},
  year = 2024,
  journal = {Remote Sensing},
  volume = {16},
  number = {13},
  pages = {2355},
  doi = {10.3390/rs16132355}
}

@article{ChangeMambaRemoteSensingChen24,
  title = {{{ChangeMamba}}: {{Remote}} Sensing Change Detection with Spatiotemporal State Space Model},
  author = {Chen, Hongruixuan and Song, Jian and Han, Chengxi and Xia, Junshi and Yokoya, Naoto},
  year = 2024,
  journal = {IEEE Transactions on Geoscience and Remote Sensing},
  volume = {62},
  pages = {1--20},
  doi = {10.1109/TGRS.2024.3417253}
}

@article{ChangeMaskDeepMultitaskZheng22,
  title = {{{ChangeMask}}: {{Deep}} Multi-Task Encoder-Transformer-Decoder Architecture for Semantic Change Detection},
  author = {Zheng, Zhuo and Zhong, Yanfei and Tian, Shiqi and Ma, Ailong and Zhang, Liangpei},
  year = 2022,
  journal = {ISPRS Journal of Photogrammetry and Remote Sensing},
  volume = {183},
  pages = {228--239},
  doi = {10.1016/j.isprsjprs.2021.10.015}
}

@article{ChangerFeatureInteractionFang23,
  title = {Changer: {{Feature Interaction}} Is {{What You Need}} for {{Change Detection}}},
  author = {Fang, Sheng and Li, Kaiyu and Li, Zhe},
  year = 2023,
  journal = {IEEE Transactions on Geoscience and Remote Sensing},
  volume = {61},
  pages = {1--11},
  doi = {10.1109/TGRS.2023.3277496}
}

@article{ChangeVectorAnalysisMalila80,
  title = {Change Vector Analysis: {{An}} Approach for Detecting Forest Changes with Landsat},
  author = {Malila, William},
  year = 1980,
  journal = {LARS Symposia}
}

@misc{CoRegOVCDConsistencyregularizedOpenvocabularyTang26,
  title = {{{CoRegOVCD}}: {{Consistency}}-Regularized Open-Vocabulary Change Detection},
  author = {Tang, Weidong and Sun, Hanbin and Li, Zihan and Wang, Yikai and Zhang, Feifan},
  year = 2026,
  number = {arXiv:2604.02160},
  eprint = {2604.02160},
  doi = {10.48550/arXiv.2604.02160}
}

@article{CrossDomainChangeDetectionSong23,
  title = {A {{Cross-Domain Change Detection Network Based}} on {{Instance Normalization}}},
  author = {Song, Yabin and Xiang, Jun and Jiang, Jiawei and Yan, Enping and Wei, Wei and Mo, Dengkui},
  year = 2023,
  journal = {Remote Sensing},
  volume = {15},
  number = {24},
  pages = {5785},
  doi = {10.3390/rs15245785}
}

@article{CSDACDDomainAdaptiveChangeHuang24,
  title = {{{CSDACD}}: {{Domain-Adaptive Change Detection Network}} for {{Cross-Seasonal Remote Sensing Images}}},
  author = {Huang, Yuxing and Zhang, Peng},
  year = 2024,
  journal = {IEEE Geoscience and Remote Sensing Letters},
  volume = {21},
  pages = {1--5},
  doi = {10.1109/LGRS.2024.3431212}
}

@article{DeeplySupervisedImageZhang20,
  title = {A Deeply Supervised Image Fusion Network for Change Detection in High Resolution Bi-Temporal Remote Sensing Images},
  author = {Zhang, Chenxiao and Yue, Peng and Tapete, Deodato and Jiang, Liangcun and Shangguan, Boyi and Huang, Li and Liu, Guangchao},
  year = 2020,
  journal = {ISPRS Journal of Photogrammetry and Remote Sensing},
  volume = {166},
  pages = {183--200},
  doi = {10.1016/j.isprsjprs.2020.06.003}
}

@inproceedings{DiffRegCDIntegratedRegistrationMadani26,
  title = {{{DiffRegCD}}: {{Integrated Registration}} and {{Change Detection}} with {{Diffusion Features}}},
  booktitle = {Proceedings of the {{IEEE}}/{{CVF Winter Conference}} on {{Applications}} of {{Computer Vision}}},
  author = {Madani, Seyedehanita and Chellappa, Rama and Patel, Vishal M.},
  year = 2026,
  pages = {7503--7512}
}

@article{DINOv2LearningRobustOquab24,
  title = {{{DINOv2}}: {{Learning Robust Visual Features}} without {{Supervision}}},
  author = {Oquab, Maxime and Darcet, Timoth{\'e}e and Moutakanni, Th{\'e}o and Vo, Huy V. and Szafraniec, Marc and Khalidov, Vasil and Fernandez, Pierre and Haziza, Daniel and Massa, Francisco and {El-Nouby}, Alaaeldin and Assran, Mido and Ballas, Nicolas and Galuba, Wojciech and Howes, Russell and Huang, Po-Yao and Li, Shang-Wen and Misra, Ishan and Rabbat, Michael and Sharma, Vasu and Synnaeve, Gabriel and Xu, Hu and J{\'e}gou, Herv{\'e} and Mairal, Julien and Labatut, Patrick and Joulin, Armand and Bojanowski, Piotr},
  year = 2024,
  journal = {Transactions on Machine Learning Research},
  volume = {2024}
}

@article{DINOv3Simeoni26,
  title = {{{DINOv3}}},
  author = {Sim{\'e}oni, Oriane and Vo, Huy V. and Seitzer, Maximilian and Baldassarre, Federico and Oquab, Maxime and Jose, Cijo and Khalidov, Vasil and Szafraniec, Marc and Yi, Seung Eun and Ramamonjisoa, Michael and Massa, Francisco and Haziza, Daniel and Wehrstedt, Luca and Wang, Jianyuan and Darcet, Timoth{\'e}e and Moutakanni, Th{\'e}o and Sentana, Leonel and Roberts, Claire and Vedaldi, Andrea and Tolan, Jamie and Brandt, John and Couprie, Camille and Mairal, Julien and Jegou, Herve and Labatut, Patrick and Bojanowski, Piotr},
  year = 2026,
  journal = {Transactions on Machine Learning Research}
}

@article{DynamicEarthHowFarLi26,
  title = {{{DynamicEarth}}: {{How Far Are We}} from {{Open-Vocabulary Change Detection}}?},
  author = {Li, Kaiyu and Cao, Xiangyong and Deng, Yupeng and Pang, Chao and Xin, Zepeng and Qiao, Hui and Gong, Tieliang and Meng, Deyu and Wang, Zhi},
  year = 2026,
  journal = {Proceedings of the AAAI Conference on Artificial Intelligence},
  volume = {40},
  number = {8},
  pages = {6279--6287},
  doi = {10.1609/aaai.v40i8.37554}
}

@article{EdgeRefNetEdgeguidedRefinementHussin26,
  title = {{{EdgeRefNet}}: An Edge-Guided Refinement Network for Building Change Detection in Remote Sensing Images},
  author = {Hussin, Wafaa I. M. and Lu, Zhi and Ashraf, Aysha and Mao, Aji and Peng, Zhenming},
  year = 2026,
  journal = {IEEE Transactions on Geoscience and Remote Sensing},
  volume = {64},
  pages = {4409610--4409610},
  doi = {10.1109/TGRS.2026.3689397}
}

@inproceedings{FewShotChangeDetectionCao25,
  author = {Cao, Zhi-juan and Liu, Jian-wei},
  title = {Few-Shot Change Detection: A General Class-Independent Change Detection Algorithm},
  booktitle = {2025 37th Chinese Control and Decision Conference (CCDC)},
  year = {2025},
  pages = {2975--2982},
  doi = {10.1109/CCDC65474.2025.11090552}
}

@article{FocusedInformationLearningAhn24,
  title = {Focused Information Learning Method for Change Detection Based on Segmentation with Limited Annotations},
  author = {Ahn, H. and Chung, S. and Park, S. and Kim, D.},
  year = 2024,
  journal = {International Journal of Applied Earth Observation and Geoinformation},
  volume = {129},
  pages = {103839},
  doi = {10.1016/j.jag.2024.103839}
}

@misc{FROSTTrainingFreePark26,
  author = {Park, Junghwan},
  title = {{FROST}: Training-Free Few-Shot Segmentation with Frozen Features and Nonparametric Statistics},
  year = {2026},
  eprint = {2606.31136},
  archivePrefix = {arXiv},
  primaryClass = {cs.CV},
  url = {https://arxiv.org/abs/2606.31136}
}

@article{FullyConvolutionalNetworksJi19,
  title = {Fully {{Convolutional Networks}} for {{Multisource Building Extraction From}} an {{Open Aerial}} and {{Satellite Imagery Data Set}}},
  author = {Ji, Shunping and Wei, Shiqing and Lu, Meng},
  year = 2019,
  journal = {IEEE Transactions on Geoscience and Remote Sensing},
  volume = {57},
  number = {1},
  pages = {574--586},
  doi = {10.1109/TGRS.2018.2858817}
}

@inproceedings{FullyConvolutionalSiameseCayeDaudt18,
  title = {Fully Convolutional Siamese Networks for Change Detection},
  booktitle = {{{IEEE International Conference}} on {{Image Processing}}},
  author = {Caye Daudt, Rodrigo and Le Saux, Bertr and Boulch, Alexandre},
  year = 2018,
  pages = {4063--4067},
  doi = {10.1109/ICIP.2018.8451652}
}

@misc{HIMECDirectionalChangeAshraf26,
  title = {{{HIMEC}}: {{Directional Change Representation}} and {{Fixed-Interface Decoding}} for {{Remote Sensing Image Change Captioning}}},
  author = {Ashraf, Aysha and Ashraf, Shaina and Hussin, Wafaa I. M. and Haider, Ali and Lu, Zhi and Peng, Zhenming},
  year = 2026,
  number = {arXiv:2608.12502},
  eprint = {2608.12502},
  primaryclass = {cs.CV},
  doi = {10.48550/arXiv.2608.12502}
}

@inproceedings{ImagesSpeakImagesWang23,
  title = {Images {{Speak}} in {{Images}}: {{A Generalist Painter}} for {{In-Context Visual Learning}}},
  booktitle = {Proceedings of the {{IEEE}}/{{CVF Conference}} on {{Computer Vision}} and {{Pattern Recognition}}},
  author = {Wang, Xinlong and Wang, Wen and Cao, Yue and Shen, Chunhua and Huang, Tiejun},
  year = 2023,
  eprint = {2212.02499},
  pages = {6830--6839},
  doi = {10.1109/CVPR52729.2023.00660}
}

@inproceedings{INSID3TrainingfreeIncontextCuttano26,
  title = {{{INSID3}}: {{Training}}-Free in-Context Segmentation with {{DINOv3}}},
  booktitle = {Proceedings of the {{IEEE}}/{{CVF Conference}} on {{Computer Vision}} and {{Pattern Recognition}}},
  author = {Cuttano, Claudia and Trivigno, Gabriele and Reich, Christoph and Cremers, Daniel and Masone, Carlo and Roth, Stefan},
  year = 2026
}

@article{LearningDetectChangesMesquita22,
  title = {Learning to {{Detect Changes}} in {{Aerial Images}} in the {{Presence}} of {{Registration Errors}}},
  author = {Mesquita, Daniel B. and Campos, Mario F. M. and Nascimento, Erickson R.},
  year = 2022,
  journal = {IEEE Geoscience and Remote Sensing Letters},
  volume = {19},
  pages = {1--5},
  doi = {10.1109/LGRS.2022.3193375}
}

@inproceedings{LearningTransferableVisualRadford21,
  title = {Learning Transferable Visual Models from Natural Language Supervision},
  booktitle = {Proceedings of the 38th {{International Conference}} on {{Machine Learning}}},
  author = {Radford, Alec and Kim, Jong Wook and Hallacy, Chris and Ramesh, Aditya and Goh, Gabriel and Agarwal, Sandhini and Sastry, Girish and Askell, Amanda and Mishkin, Pamela and Clark, Jack and Krueger, Gretchen and Sutskever, Ilya},
  year = 2021,
  pages = {8748--8763}
}

@inproceedings{MatcherSegmentAnythingLiu23,
  title = {Matcher: {{Segment Anything}} with {{One Shot Using All-Purpose Feature Matching}}},
  booktitle = {{{ICLR}}},
  author = {Liu, Yang and Zhu, Muzhi and Li, Hengtao and Chen, Hao and Wang, Xinlong and Shen, Chunhua},
  year = 2023
}

@misc{MemOVCDTrainingFreeOpenVocabularyKuang26,
  title = {{{MemOVCD}}: {{Training-Free Open-Vocabulary Change Detection}} via {{Cross-Temporal Memory Reasoning}} and {{Global-Local Adaptive Rectification}}},
  author = {Kuang, Zuzheng and Chang, Honghao and Liang, Boqiang and Wang, Haoqian and He, Lijun and Li, Fan and Bi, Haixia},
  year = 2026,
  number = {arXiv:2604.26774},
  eprint = {2604.26774},
  primaryclass = {cs.CV},
  doi = {10.48550/arXiv.2604.26774}
}

@inproceedings{MergeSAMUnsupervisedChangeHu25,
  title = {{{MergeSAM}}: {{Unsupervised Change Detection Of Remote Sensing Images Based}} on the {{Segment Anything Model}}},
  booktitle = {{{IEEE International Geoscience}} and {{Remote Sensing Symposium}}},
  author = {Hu, Meiqi and Lu, Lingzhi and Han, Chengxi and Liu, Xiaoping},
  year = 2025,
  pages = {8116--8119},
  doi = {10.1109/IGARSS55030.2025.11243146}
}

@article{OCAObjectbasedChangeZhang25,
  title = {{{OCA}}: {{Object}}-Based Change Augmentation for Few-Shot Building Change Detection in Very High-Resolution Remote Sensing Images},
  author = {Zhang, Chenxiao and Yue, Peng and Cigna, Francesca and Tapete, Deodato},
  year = 2025,
  journal = {IEEE Transactions on Geoscience and Remote Sensing},
  volume = {63},
  pages = {1--14},
  doi = {10.1109/TGRS.2025.3584092}
}

@misc{OmniOVCDStreamliningOpenVocabularyZhang26,
  title = {{{OmniOVCD}}: {{Streamlining Open-Vocabulary Change Detection}} with {{SAM}} 3},
  author = {Zhang, Xu and Li, Danyang and Xia, Yingjie and Dong, Xiaohang and Yu, Hualong and Wang, Jianye and Li, Qicheng},
  year = 2026,
  number = {arXiv:2601.13895},
  eprint = {2601.13895},
  primaryclass = {cs},
  doi = {10.48550/arXiv.2601.13895}
}

@inproceedings{OpenDPROpenVocabularyChangeGuo26,
  title = {{{OpenDPR}}: {{Open-Vocabulary Change Detection}} via {{Vision-Centric Diffusion-Guided Prototype Retrieval}} for {{Remote Sensing Imagery}}},
  booktitle = {Proceedings of the {{IEEE}}/{{CVF Conference}} on {{Computer Vision}} and {{Pattern Recognition}}},
  author = {Guo, Qi and Wang, Jue and Liu, Yinhe and Zhong, Yanfei},
  year = 2026,
  pages = {20399--20409}
}

@article{OvercomingUncertaintyChallengesLi25,
  title = {Overcoming the Uncertainty Challenges in Detecting Building Changes from Remote Sensing Images},
  author = {Li, Jiepan and He, Wei and Li, Zhuohong and Guo, Yujun and Zhang, Hongyan},
  year = 2025,
  journal = {ISPRS Journal of Photogrammetry and Remote Sensing},
  volume = {220},
  pages = {1--17},
  doi = {10.1016/j.isprsjprs.2024.11.017}
}

@inproceedings{PersonalizeSegmentAnythingZhang24,
  title = {Personalize {{Segment Anything Model}} with {{One Shot}}},
  booktitle = {The {{Twelfth International Conference}} on {{Learning Representations}}},
  author = {Zhang, Renrui and Jiang, Zhengkai and Guo, Ziyu and Yan, Shilin and Pan, Junting and Dong, Hao and Qiao, Yu and Gao, Peng and Li, Hongsheng},
  year = 2024
}

@article{PFENetPriorGuidedTian22,
  author = {Tian, Zhuotao and Zhao, Hengshuang and Shu, Michelle and Yang, Zhicheng and Li, Ruiyu and Jia, Jiaya},
  title = {Prior Guided Feature Enrichment Network for Few-Shot Segmentation},
  journal = {IEEE Transactions on Pattern Analysis and Machine Intelligence},
  year = {2022},
  volume = {44},
  number = {2},
  pages = {1050--1065},
  doi = {10.1109/TPAMI.2020.3013717}
}

@article{PostclassificationChangeDetectionWu17,
  title = {A Post-Classification Change Detection Method Based on Iterative Slow Feature Analysis and {{Bayesian}} Soft Fusion},
  author = {Wu, Chen and Du, Bo and Cui, Xiaohui and Zhang, Liangpei},
  year = 2017,
  journal = {Remote Sensing of Environment},
  volume = {199},
  pages = {241--255},
  doi = {10.1016/j.rse.2017.07.009}
}

@misc{ReAOVCDTrainingFreeOpenVocabularyZhu26,
  title = {{{ReA-OVCD}}: {{Training-Free Open-Vocabulary Change Detection}} via {{Semantic-Spatial Reliability Assessment}}},
  author = {Zhu, Hongming and Chen, Huaji and Du, Bowen and Liu, Sicong and Liu, Qin},
  year = 2026,
  number = {arXiv:2606.20032},
  eprint = {2606.20032},
  primaryclass = {cs.CV},
  doi = {10.48550/arXiv.2606.20032}
}

@misc{REBASEReferenceBackgroundSubspaceGopal26,
  title = {{{REBASE}}: {{Reference-Background Subspace Elimination}} for {{Training-Free In-Context Segmentation}}},
  author = {Gopal, Mantha Sai and Chacko, Jaison Saji and Nandwana, Harsh and Hegde, Sandesh and Banerjee, Debarshi and Mahesh, Uma},
  year = 2026,
  number = {arXiv:2607.09082},
  eprint = {2607.09082},
  primaryclass = {cs.CV},
  doi = {10.48550/arXiv.2607.09082}
}

@article{RegularizedIterativelyReweightedNielsen07,
  title = {The Regularized Iteratively Reweighted {{MAD}} Method for Change Detection in Multi- and Hyperspectral Data},
  author = {Nielsen, Allan Aasbjerg},
  year = 2007,
  journal = {IEEE Transactions on Image Processing},
  volume = {16},
  number = {2},
  pages = {463--478},
  doi = {10.1109/TIP.2006.888195}
}

@article{RemoteCLIPVisionLanguageLiu24,
  title = {{{RemoteCLIP}}: {{A Vision Language Foundation Model}} for {{Remote Sensing}}},
  author = {Liu, Fan and Chen, Delong and Guan, Zhangqingyun and Zhou, Xiaocong and Zhu, Jiale and Ye, Qiaolin and Fu, Liyong and Zhou, Jun},
  year = 2024,
  journal = {IEEE Transactions on Geoscience and Remote Sensing},
  volume = {62},
  pages = {1--16},
  doi = {10.1109/TGRS.2024.3390838}
}

@article{RemoteSensingImageChen22,
  title = {Remote Sensing Image Change Detection with Transformers},
  author = {Chen, Hao and Qi, Zipeng and Shi, Zhenwei},
  year = 2022,
  journal = {IEEE Transactions on Geoscience and Remote Sensing},
  volume = {60},
  eprint = {2103.00208},
  primaryclass = {cs},
  pages = {1--14},
  doi = {10.1109/TGRS.2021.3095166}
}

@article{S2LookingSatelliteSidelookingShen21,
  title = {{{S2Looking}}: A Satellite Side-Looking Dataset for Building Change Detection},
  author = {Shen, Li and Lu, Yao and Chen, Hao and Wei, Hao and Xie, Donghai and Yue, Jiabao and Chen, Rui and Lv, Shouye and Jiang, Bitao},
  year = 2021,
  journal = {Remote Sensing},
  volume = {13},
  number = {24},
  pages = {5094},
  doi = {10.3390/rs13245094}
}

@misc{Seg2ChangeAdaptingOpenVocabularySu26,
  title = {{{Seg2Change}}: {{Adapting Open-Vocabulary Semantic Segmentation Model}} for {{Remote Sensing Change Detection}}},
  author = {Su, You and Song, Yonghong and Chen, Jingqi and Wen, Zehan},
  year = 2026,
  number = {arXiv:2604.11231},
  eprint = {2604.11231},
  primaryclass = {cs.CV},
  doi = {10.48550/arXiv.2604.11231}
}

@inproceedings{SegGPTSegmentingWang23,
  author = {Wang, Xinlong and Zhang, Xiaosong and Cao, Yue and Wang, Wen and Shen, Chunhua and Huang, Tiejun},
  title = {{SegGPT}: Towards Segmenting Everything in Context},
  booktitle = {Proceedings of the {IEEE}/{CVF} International Conference on Computer Vision},
  year = {2023},
  pages = {1130--1140},
  doi = {10.1109/ICCV51070.2023.00110}
}

@inproceedings{SegmentAnyChangeZheng24,
  title = {Segment Any Change},
  booktitle = {Advances in {{Neural Information Processing Systems}}},
  author = {Zheng, Zhuo and Zhong, Yanfei and Zhang, Liangpei and Ermon, Stefano},
  year = 2024,
  volume = {37},
  pages = {81204--81224},
  doi = {10.52202/079017-2581}
}

@inproceedings{SegmentAnythingKirillov23,
  title = {Segment Anything},
  booktitle = {2023 {{IEEE}}/{{CVF International Conference}} on {{Computer Vision}} ({{ICCV}})},
  author = {Kirillov, Alexander and Mintun, Eric and Ravi, Nikhila and Mao, Hanzi and Rolland, Chloe and Gustafson, Laura and Xiao, Tete and Whitehead, Spencer and Berg, Alexander C. and Lo, Wan-Yen and Doll{\'a}r, Piotr and Girshick, Ross},
  year = 2023,
  pages = {3992--4003},
  doi = {10.1109/ICCV51070.2023.00371}
}

@inproceedings{SegmentChangeModelTan24,
  title = {Segment Change Model ({{SCM}}) for Unsupervised Change Detection in {{VHR}} Remote Sensing Images: A Case Study of Buildings},
  booktitle = {{{IEEE International Geoscience}} and {{Remote Sensing Symposium}}},
  author = {Tan, Xiaoliang and Chen, Guanzhou and Wang, Tong and Wang, Jiaqi and Zhang, Xiaodong},
  year = 2024,
  pages = {8577--8580},
  doi = {10.1109/IGARSS53475.2024.10642429}
}

@misc{SemanticSelectionGapZakir26,
  author = {Zakir, Hussni Mohd and Ho, Eric Tatt Wei},
  title = {Revealing the Semantic Selection Gap in {DINOv3} through Training-Free Few-Shot Segmentation},
  year = {2026},
  eprint = {2602.07550},
  archivePrefix = {arXiv},
  primaryClass = {cs.CV},
  url = {https://arxiv.org/abs/2602.07550}
}

@article{SNUNetCDDenselyConnectedFang22,
  title = {{{SNUNet-CD}}: A Densely Connected Siamese Network for Change Detection of {{VHR}} Images},
  author = {Fang, Sheng and Li, Kaiyu and Shao, Jinyuan and Li, Zhe},
  year = 2022,
  journal = {IEEE Geoscience and Remote Sensing Letters},
  volume = {19},
  pages = {1--5},
  doi = {10.1109/LGRS.2021.3056416}
}

@misc{SpatiallySelectiveSelfTrainingHussin26,
  title = {Spatially {{Selective Self-Training}} for {{Unsupervised Building Change Detection}}},
  author = {Hussin, Wafaa I. M. and Lu, Zhi and Mohammed, Anas M. I. and Zhou, Xiang and Abubaker, Ratiba A. H. and Peng, Zhenming},
  year = 2026,
  number = {arXiv:2606.10775},
  eprint = {2606.10775},
  primaryclass = {cs.CV},
  doi = {10.48550/arXiv.2606.10775}
}

@article{SpatialTemporalAttentionBasedMethodChen20,
  title = {A {{Spatial-Temporal Attention-Based Method}} and a {{New Dataset}} for {{Remote Sensing Image Change Detection}}},
  author = {Chen, Hao and Shi, Zhenwei},
  year = 2020,
  journal = {Remote Sensing},
  volume = {12},
  number = {10},
  pages = {1662},
  doi = {10.3390/rs12101662}
}

@inproceedings{TransformerbasedSiameseNetworkBandara22,
  title = {A Transformer-Based Siamese Network for Change Detection},
  booktitle = {{{IEEE International Geoscience}} and {{Remote Sensing Symposium}}},
  author = {Bandara, Wele Gedara Chaminda and Patel, Vishal M.},
  year = 2022,
  pages = {207--210},
  doi = {10.1109/IGARSS46834.2022.9883686}
}

@misc{UniVCDNewMethodZhu25,
  title = {{{UniVCD}}: A New Method for Unsupervised Change Detection in the Open-Vocabulary Era},
  author = {Zhu, Ziqiang and Yang, Bowei},
  year = 2025,
  number = {arXiv:2512.13089},
  eprint = {2512.13089},
  doi = {10.48550/arXiv.2512.13089}
}

@article{UnsupervisedChangeDetectionCelik09,
  title = {Unsupervised Change Detection in Satellite Images Using Principal Component Analysis and K-Means Clustering},
  author = {Celik, Turgay},
  year = 2009,
  journal = {IEEE Geoscience and Remote Sensing Letters},
  volume = {6},
  number = {4},
  pages = {772--776},
  doi = {10.1109/LGRS.2009.2025059}
}

@article{UnsupervisedChangeDetectionNoh24,
  title = {Unsupervised Change Detection Based on Image Reconstruction Loss with Segment Anything},
  author = {Noh, HyeonCheol and Ju, JinGi and Kim, YuHyun and Kim, MinWoo and Choi, Dong-Geol},
  year = 2024,
  journal = {Remote Sensing Letters},
  volume = {15},
  number = {9},
  pages = {919--929},
  doi = {10.1080/2150704X.2024.2388851}
}

@article{UnsupervisedDeepChangeSaha19,
  title = {Unsupervised Deep Change Vector Analysis for Multiple-Change Detection in {{VHR}} Images},
  author = {Saha, Sudipan and Bovolo, Francesca and Bruzzone, Lorenzo},
  year = 2019,
  journal = {IEEE Transactions on Geoscience and Remote Sensing},
  volume = {57},
  number = {6},
  pages = {3677--3693},
  doi = {10.1109/TGRS.2018.2886643}
}

@article{UnsupervisedDeepSlowDu19,
  title = {Unsupervised {{Deep Slow Feature Analysis}} for {{Change Detection}} in {{Multi-Temporal Remote Sensing Images}}},
  author = {Du, Bo and Ru, Lixiang and Wu, Chen and Zhang, Liangpei},
  year = 2019,
  journal = {IEEE Transactions on Geoscience and Remote Sensing},
  volume = {57},
  number = {12},
  pages = {9976--9992},
  doi = {10.1109/TGRS.2019.2930682}
}
}

\end{document}